\documentclass[10pt,letterpaper]{article}
\usepackage[top=0.85in,left=2.75in,footskip=0.75in]{geometry}

\usepackage{amsmath,amssymb}

\usepackage{changepage}

\usepackage[utf8x]{inputenc}

\usepackage{textcomp,marvosym}

\usepackage{cite}

\usepackage{nameref,hyperref}

\usepackage{microtype}
\DisableLigatures[f]{encoding = *, family = * }

\usepackage[table]{xcolor}

\usepackage{array}

\usepackage{wrapfig}
\usepackage{hyperref}
\usepackage{microtype}
\usepackage{graphicx}
\usepackage{subfigure}
\usepackage{booktabs} 
\usepackage{amsmath}
\usepackage{centernot}
\usepackage{amssymb}

\newcolumntype{+}{!{\vrule width 2pt}}

\newlength\savedwidth

\raggedright
\usepackage[aboveskip=1pt,labelfont=bf,labelsep=period,justification=raggedright,singlelinecheck=off]{caption}

\makeatletter
\renewcommand{\@biblabel}[1]{\quad#1.}
\makeatother

\usepackage{lastpage,fancyhdr,graphicx}
\usepackage{epstopdf}
\fancyheadoffset[L]{2.25in}
\fancyfootoffset[L]{2.25in}
\begin{document}
\vspace*{0.2in}

\begin{flushleft}
{\Large
\textbf\newline{Semi-supervised Contrastive Learning Using Partial Label Information} 
}
\newline
\\
Colin B. Hansen*\textsuperscript{1},
Vishwesh Nath\textsuperscript{1},
Diego A. Mesa\textsuperscript{2},
Yuankai Huo\textsuperscript{3},
Bennett A. Landman\textsuperscript{1,3},
Thomas A. Lasko\textsuperscript{2}
\\
\bigskip
\textbf{1} Department of Computer Science, Vanderbilt University, Nashville, TN, USA
\\
\textbf{2} Department of Biomedical Informatics, Vanderbilt University Medical Center, Nashville, TN, USA
\\
\textbf{3} Department of Electrical Engineering, Vanderbilt University Medical Center, Nashville, TN, USA
\\
\bigskip

%
%





* colin.b.hansen@vanderbilt.edu

\end{flushleft}
\section*{Abstract}
 In semi-supervised learning, information from unlabeled examples is used to improve the model learned from labeled examples. In some learning problems, partial label information can be inferred from otherwise unlabeled examples and used to further improve the model.  In particular, partial label information exists when subsets of training examples are known to have the same label, even though the label itself is missing. By encouraging the model to give the same label to all such examples through contrastive learning objectives, we can potentially improve its performance. We call this encouragement  \emph{Nullspace Tuning} because the difference vector between any pair of examples with the same label should lie in the nullspace of a linear model.  In this paper, we investigate the benefit of using partial label information using a careful comparison framework over well-characterized public datasets. We show that the additional information provided by partial labels reduces test error over good semi-supervised methods usually by a factor of 2, up to a factor of 5.5 in the best case. We also show that adding Nullspace Tuning to the newer and state-of-the-art MixMatch method decreases its test error by up to a factor of 1.8.


\section*{Introduction}


Semi-supervised learning methods attempt to improve a model learned from labeled examples by using information extracted from unlabeled examples \cite{Chapelle2006, zhu2003semi, verma2019interpolation}. One effective approach is to use some form of data augmentation, in which a new example is created by transforming an unlabeled example \cite{Cubuk2019,berthelot2019mixmatch}, and then encouraging the model to predict the same label for both.

\begin{figure*}[h!]
\vskip 0.2in
\begin{center}
\centerline{\includegraphics[width=12cm]{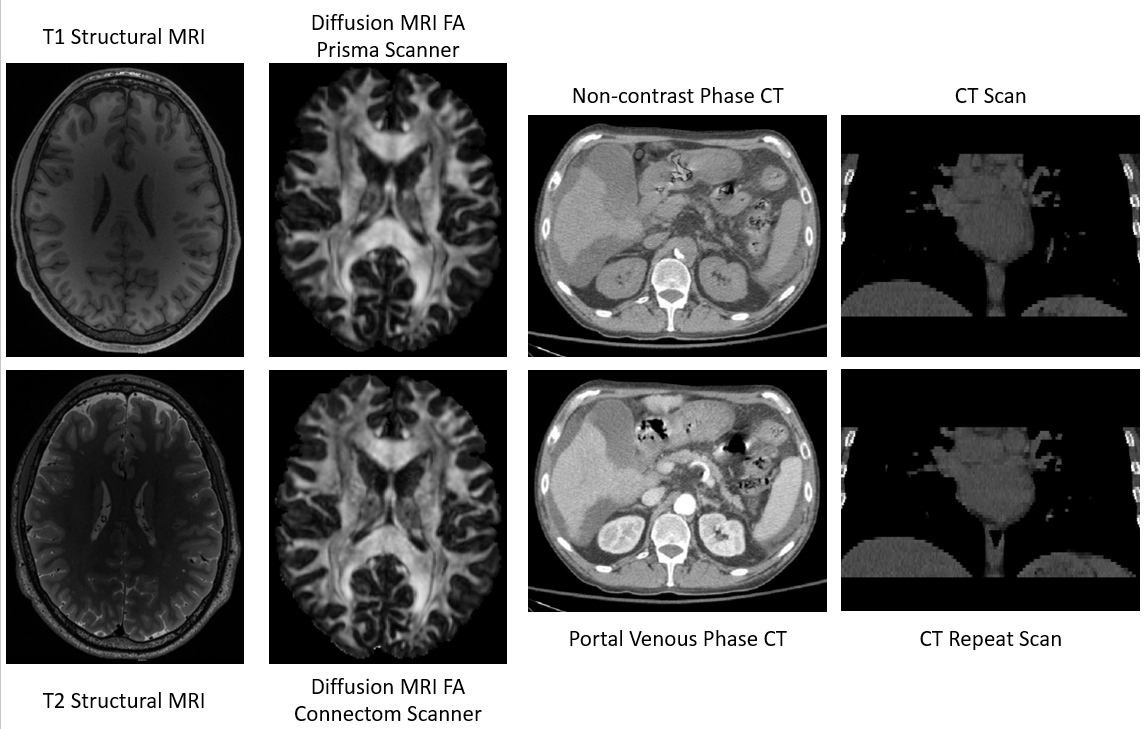}}
\caption{In medical imaging repeat partial label information commonly comes in the form of repeat acquisitions of a subject. Assuming these acquisitions are acquired within a reasonable amount of time such that aging does not affect the anatomy, models can leverage the differences between acquisitions that may arise from differences in acquisition parameters. This may be differences in contrast such as the difference between T1 weighted MRI (top left) and T2 MRI (bottom left) or between non-contrast phase CT (top center-right) and portal venous phase CT (bottom center-right). The manufacturer of the imaging equipment may be a factor as well as is shown in the diffusion MRI fractional anisotropy (FA) estimated from a Prisma scanner (top center-left) and the FA estimated from a Connectom scanner (bottom center-left). Using repeat acquisitions with the same parameters and hardware can also provide useful information such as in repeat heart CT (right). }
\label{equiv_examples}
\end{center}
\vskip -0.2in
\end{figure*}

In some learning problems, we already know that a fraction of unlabeled examples have the same label, even though that label is missing. For example, we may know that multiple photos are of the same object because of the way the photos were acquired, despite not having a label for that object. In the medical domain, repeated imaging of the same patient is common \cite{yankelevitz1999small,freeborough1997boundary,dirix2009dose}, and if the learning task is to predict something that does not change over time (or at least not over the short time between images), then we may know that the repeated images have the same label depending on the domain (Fig \ref{equiv_examples}). We call this knowledge \emph{partial label information}, and distinguish it from the standard semi-supervised assumption that there is no label information at all for the unlabeled examples. 

In prior work, Nath \cite{nath2018inter} used partial label information to predict fiber orientation distributions in magnetic resonance imaging, and  Huo \cite{huo2019coronary} used it for coronary artery calcium detection in non-contrast computed tomography scans. But there has been no careful investigation of how much the partial label information can add to a model's performance. In this paper, we evaluate the performance benefit of partial labels using the rigorous, standardized approach described by Oliver \cite{oliver2018realistic}, which is designed to realistically assess the relative performance of semi-supervised learning approaches.

In this work, we use the term \emph{equivalence class} to indicate a subset of unlabeled examples for which the label is known to be the same. An equivalence class need not contain \emph{all} of the examples with the same label. Formally, an equivalence class $Q$ of examples $x$ in a data subset $D$ under a true but unknown labeling function $f$ is defined as:

\begin{equation}
\label{EQ_definition}
    Q = \{x \in D | f(x) = c\},
\end{equation}

where $c$ is a constant. If we know that a particular pair $x_1, x_2 \in Q$, we express this as $x_1 \sim x_2$.

 If the labeling function $f$ is a linear operator, then the difference between any two examples in $Q$ lies in the nullspace of $f$: 
\begin{equation}
\label{linear_nullspace}
    f(x_1) = f(x_2) \iff f(x_1 - x_2) = 0. 
\end{equation}

Because the term \emph{nullspace} is so evocative, we abuse it here to conceptually refer to comparisons between elements of the sets defined by (\ref{EQ_definition}), even though for a nonlinear function $f$ the relationship (\ref{linear_nullspace}) does not hold.

We can use our knowledge of the natural equivalence classes in a dataset to help tune a model using a procedure that we call \emph{Nullspace Tuning}, in which the model is encouraged to label examples $x_1$ and $x_2$ the same when $x_1 \sim x_2$. Nullspace tuning is easily implemented by adding a term to the loss function to penalize the difference in label probabilities $h(x_1) - h(x_2)$ assigned by the current model $h$ when $x_1 \sim x_2$.

Nullspace Tuning is related to the idea of Data Augmentation (see Section \ref{sec:data-augmentation} below) and is a form of contrastive learning. Data Augmentation is used when we know specific transformations that should not affect the label, and we create new training examples by transforming existing examples and keeping the label constant, whether that label is known or unknown. This \emph{implicitly} places those transformations into the architecture's nullspace. Contrastive Learning constrains a network using two samples from the same or different classes. Some semi-supervised contrastive methods create two instances from the same sample using data augmentation \cite{chen2020simple}. In contrast, Nullspace Tuning \emph{explicitly} places naturally occurring but unknown transformations into the nullspace when nature provides examples of them by way of partial labels. The contribution of this work is the empirical analysis of the use of partial label information to improve model generalizability through contrastive learning. In specific cases, unlabeled data may contain partial label information in the form of equivalence classes which Nullspace Tuning proposes to use to identify non-class altering differences between two samples.

\section*{Related Work}

 Data augmentation, semi-supervised learning, and contrastive learning are existing learning approaches that are closely related to Nullspace Tuning, and there is a large literature for each. Thorough reviews are available elsewhere \cite{perez2017effectiveness,zhu2009introduction,zhu2005semi,kumar2016learning}. In this section we discuss specific work that is most closely related, and work that we use as experimental baselines. 

\subsection*{Data Augmentation}
\label{sec:data-augmentation}
Data Augmentation artificially expands a training dataset by modifying examples using transformations that are believed not to affect the label. Image deformations and additive noise are common examples of such transformations \cite{cirecsan2010deep,Cubuk2019,simard2003best}. The most effective data augmentations may be specific to the learning task or dataset and driven by domain knowledge. Elastic distortions, scale, translation, and rotation are used in the majority of top performing MNIST models \cite{wan2013regularization,sato2015apac,simard2003best,ciregan2012multi}. Random cropping, mirroring, and color shifting are often used to augment natural images \cite{krizhevsky2012imagenet}. Recent work automatically selects effective data augmentation policies from a search space of image processing functions \cite{Cubuk2019}. Though data augmentation can provide useful variation which enable more generalizability, they only approximate the natural differences within a class which can impact a model. When partial label information is available where the labels themselves are not, the contrast between equivalent samples is more informative than artificial augmentations. 

\subsection*{Semi-supervised Learning}
Recent approaches to Semi-supervised Learning add to the loss function a term computed over unlabeled data that encourages the model to generalize more effectively. While there are many examples of this approach \cite{oliver2018realistic}, we describe those here that we use for comparison in our experiments.

$\Pi$-Model encourages consistency between multiple predictions of the same example under the perturbations of data augmentation or dropout. The loss term penalizes the distance between the model's prediction of two perturbations of the same sample \cite{laine2016temporal,sajjadi2016regularization}. 

Mean Teacher  \cite{tarvainen2017mean} builds on $\Pi$-Model by stabilizing the target for unlabeled samples. The target for unlabeled samples is generated from a teacher model using the exponential moving average of the student model's weights. This allows information to be aggregated after every step rather than after every epoch. 

Virtual Adversarial Training (VAT) approximates a tiny perturbation which, if added to unlabeled sample $x$, would most significantly change the resulting prediction without altering the underlying class \cite{miyato2018virtual}. VAT can be used in place of or in addition to data augmentation.

Pseudo-labeling uses the prediction function to repeatedly update the class probabilities for an unlabeled sample during training \cite{lee2013pseudo}. Probabilities that are higher than a selected threshold are treated as targets in the loss function, but typically the unlabeled portion of the loss is regulated by another hyperparameter \cite{lee2013pseudo}.

MixUp creates augmented data by forming linear interpolations between examples. If the two source examples have different labels, the new label is an interpolation of the two \cite{zhang2017mixup}.

MixMatch was developed by taking key aspects of dominant semi-supervised methods and incorporating them in to a single algorithm. The key steps are augmenting all examples, guessing low-entropy labels for  unlabeled data, and then applying MixUp to provide more interpolated examples between labeled, unlabeled, and augmeted data (using the guessed labels for unlabeled data) \cite{berthelot2019mixmatch}.

Berthelot et al. \cite{berthelot2019mixmatch} compares these semi-supervised methods to their proposed MixMatch method. They evaluate these on the CIFAR-10 dataset \cite{krizhevsky2009learning} and on the Street View House Number (SVHN) dataset \cite{netzer2011reading} as they simulate labeled and unlabeled data. They split the training set such that the models are trained at 250, 500, 1,000, 2,000, and 4,000 labeled data with the remaining treated as unlabeled data each time. In SVHN dataset, they find that MixUp generally has the worse performance reaching a 40\% test error with 250 labeled data while MixMatch has the best performance staying below 4\% test error at 250 labeled data. MixMatch also shows superior performance in the CIFAR-10 dataset achieving 11\% test error at 250 labeled data where the next best performing method VAT ahieves 36\% test error. While these semi-supervised approaches can incorporate unlabeled data, they rely on artificial data augmentation or on the model’s prediction function. While these can be important sources of information, they would ignore partial label information present in unlabeled data.

\subsection*{Equivalence Classes in Labeled Data}
An idea similar to Nullspace Tuning was used by Bromley \cite{bromley1994signature} in fully supervised learning, where $x_1 \sim x_2$ is known because their labels are observed. They used this fact to improve a signature verification model by minimizing distance between different signatures from the same person, essentially tuning the nullspace of the network with labeled equivalence classes. Contrastive loss extends this concept to learn from the contrast of two samples whether they are from the same or different classes \cite{chopra2005learning}\cite{hadsell2006dimensionality}. This idea inspired triplet networks \cite{schroff2015facenet} that learn from tuples $(x, x^+, x^-)$, where $x \sim x^+$ and $x \nsim x^-$, and the predicted probabilities are encouraged in the loss function to be respectively near or far. There are multiple works that indicate usage of Siamese networks for person re-identification \cite{mclaughlin2016recurrent,chung2017two,varior2016gated}. Nullspace Tuning extends these ideas to the case where the labels are missing but still known to be the same.

Semi-supervised contrastive learning uses data augmentation with a contrastive objective. Here $x^+$ can be generated from $x$ using some augmentation function \cite{chen2020simple}. Similarly, semi-supervised null space learning uses a positive and negative sample, but this technique uses two samples of the same object. For person re-identification, Zhang et al. relies on learning the null Foley-Sammon transform (NFST) \cite{guo2006null} from a labeled set and then using the model’s current prediction function alongside a nearest neighbors clustering to estimate groups of images which each consist of a single individual to create positive and negative samples \cite{zhang2016learning}. The goal of NFST and contrastive learning is to learn an embedding that satisfies zero within-class scatter and positive between-class scatter. The main difference between semi-supervised contrastive learning and semi-supervised null space learning is the use of two real samples in the same class rather than augmentations of a sample. This work extends the idea of null space learning to deep learning classification models while focusing on the specific case of contrastive learning with partial label information.

\section*{Methods}
This section describes the method of Nullspace Tuning using partial labels. We describe it first as a standalone approach, and then to illustrate how it can be combined with existing methods, illustrated it in combination with MixMatch. 

\subsection*{Null Space Tuning}
Given a set of labeled data $\{x_i, y_i\} \in D$ and unlabeled data $\{x^*_i\} \in D^*$ for which some equivalence classes are known, we perform Nullspace Tuning by adding to a standard loss function $\mathcal{L}_s$ a penalty on the difference in the predicted probabilities for pairs of elements of $D^*$. The new loss function $\mathcal{L}$ becomes
\begin{equation}
    \mathcal{L}=\mathcal{L}_s(h(x_i), y_i) + \lambda||h(x^*_j)-h(x^*_k)||_2^2
\end{equation}
where $h$ is the vector-valued prediction function of the model, $\lambda$ is a hyperparameter weighting the contribution of the nullspace loss term, and $x_j^*$ and $x_k^*$ are two unlabeled samples such that $x^*_j \sim x^*_k$ is known. Note that $x^*_j$ and $x^*_k$ have no required relationship to the labeled $x_i$. By minimizing the distance between equivalent unlabeled data $x_j^*$ and $x_k^*$, we encourage the model towards zero within-class scatter similar to null space learning [Zhang 2016], but rely on the supervised term to ensure positive between class scatter. In the first experiment, we use cross entropy as the standard loss function component $\mathcal{L}_s$.

\subsection*{MixMatchNST}
In the second experiment, we modify the MixMatch loss function with a Nullspace Tuning term, and denote this model MixMatchNST. In brief, MixMatch assigns a guessed label $\bar{q}$ to each unlabeled example $x^*$ by averaging the model's predicted class distributions across $K$ augmentations of $x^*$:
\begin{equation}
    \label{eq:assigned_label}
    \bar{q}_j=\frac{1}{K}\sum_{k=1}^{K} h(x_{j,k}^*)
\end{equation}
Temperature sharpening is then applied to the probability distribution of guessed labels to lower the entropy of those predictions for each example:
\begin{equation}
    \label{eq:sharpening}
    q_j=\frac{\bar{q}_j^{\frac{1}{T}}}{\sum_{i=1}^{L}\bar{q}_i^{\frac{1}{T}}} 
\end{equation}
where $T$ is a hyperparameter which is chosen to be $T=0.5$ as per Goodfellow et al. \cite{goodfellow2016deep}. Mixup \cite{zhang2017mixup} is then applied to the labeled data $\{x_i, y_i\}$ and unlabeled data $\{x^*_j, q_j\}$ to produce interpolated data $\{\tilde{x}_i, \tilde{y}_i\}$ and $\{\tilde{x}^*_j, \tilde{q}_j\}$. For a pair of two examples with their corresponding label probabilities $(x_1, p_1), (x_2, p_2)$, MixUp computes $(\tilde{x}, \tilde{p})$ as:
\begin{equation}
    \label{eq:mixup_beta}
    \lambda~Beta(\alpha,\alpha)
\end{equation}
\begin{equation}
    \label{eq:mixup_lambda}
    \lambda'=max(\lambda,1-\lambda)
\end{equation}
\begin{equation}
    \label{eq:mixup_input}
    \tilde{x}=\lambda'x_1+(1-\lambda')x_2
\end{equation}
\begin{equation}
    \label{eq:mixup_label}
    \tilde{p}=\lambda'p_1+(1-\lambda')p_2
\end{equation}
where $\alpha$ is a hyperparameter which is chosen to be $\alpha=0.75$ as per Goodfellow et al. \cite{goodfellow2016deep}. For each labeled sample and each unlabeled sample being $x_1$, another sample, labeled or unlabeled within the batch, is randomly selected as $x_2$ for MixUp which will result in $\{\tilde{x}_i,\tilde{y}_i\}$ and $\{\tilde{x}_i^*,\tilde{q}_i\}$. Weight decay is used during training to prevent overfitting \cite{loshchilov2018fixing}\cite{zhang2018three}. 

With the addition of Nullspace Tuning, the loss function for MixMatchNST becomes a combination of terms: the loss term $\mathcal{L}_X$ for labeled data, which in this case is the cross-entropy loss $\mathcal{L}_s$, the MixMatch loss term $\mathcal{L}_U$ for unlabeled data and guessed labels, and the Nullspace Tuning loss term $\mathcal{L}_E$:
\begin{equation}
\label{eq:MixMatch_labeled_term}
    \mathcal{L}_X=\mathcal{L}_s(h(\tilde{x}_i),\tilde{y}_i)
\end{equation}
\begin{equation}
\label{eq:MixMatch_unlabeled_term}
    \mathcal{L}_U = ||h(\tilde{x}^*_j)-\tilde{q}_j)||_2^2 
\end{equation}
\begin{equation}
\label{eq:NST_term}
    \mathcal{L}_E = ||q_j-q_k||_2^2
\end{equation}
\begin{equation}
    \mathcal{L} = \mathcal{L}_X + \lambda_U \mathcal{L}_U + \lambda_E \mathcal{L}_E
\end{equation}
where $\lambda_U$ and $\lambda_E$ are hyperparameters controlling the balance of terms, and $x^*_k$ is chosen so that $x^*_j \sim x^*_k$ . The added Nullspace Tuning term (\ref{eq:NST_term}) is calculated between the guessed labels $q_j$ and $q_k$ before the MixUp step, whereas the MixMatch terms (\ref{eq:MixMatch_labeled_term}) and (\ref{eq:MixMatch_unlabeled_term}) are calculated using interpolated, post-MixUp examples, as usual. 

\section*{Experiments}
We evaluate the benefit of Nullspace Tuning over partial label information using standard benchmark datasets. We follow the precedent of simulating randomly unlabeled data in these datasets \cite{oliver2018realistic}\cite{berthelot2019mixmatch}, and we likewise simulate partial labels and their equivalence classes.

\subsection*{Implementation details}
All experiments use a ``Wide ResNet-28" model \cite{zagoruyko2016paying}, with modifications made to the loss function as needed to instantiate the various comparison methods. The training procedure and error reporting follows Oliver \cite{oliver2018realistic} for our experiment comparing standalone semi-supervised methods, and Berthelot \cite{berthelot2019mixmatch} for our experiment comparing combined methods.

\subsubsection*{Standalone Methods}
\begin{figure*}[h!]
\vskip 0.2in
\begin{center}
\centerline{\includegraphics[width=14cm]{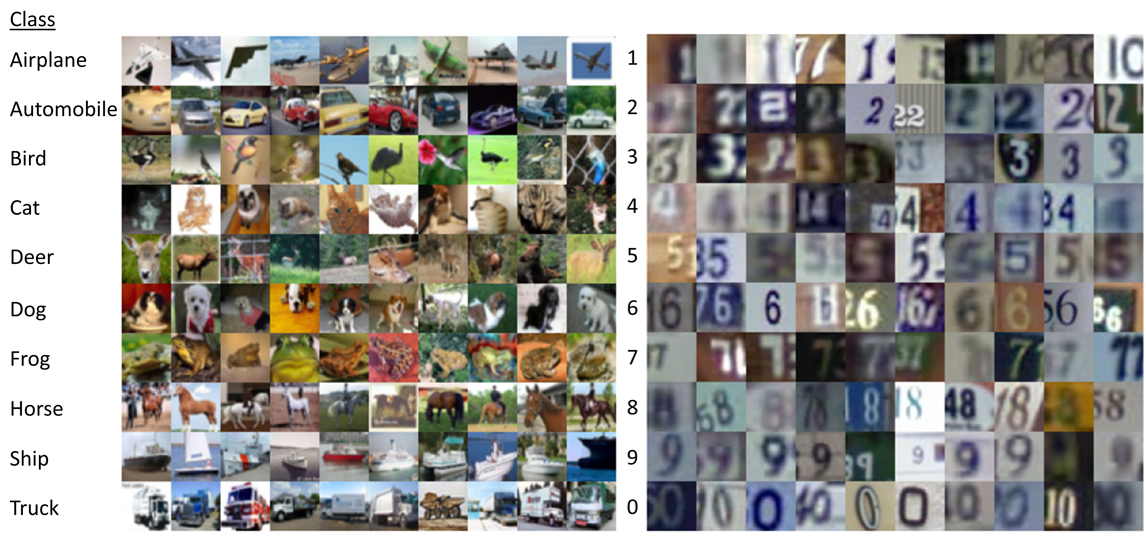}}
\caption{Samples from CIFAR-10 (left) and SVHN (right) are shown here. CIFAR-10 contains natural images of animals and vehicles and SVHN contains natural images of house numbers where the centered number is the one of interest. }
\label{cifar_svhn}
\end{center}
\vskip -0.2in
\end{figure*}

In this experiment, the simple use of Nullspace Tuning over partial labels was compared against four good  semi-supervised learning methods, using benchmark datasets CIFAR-10 \cite{krizhevsky2009learning} and SVHN \cite{netzer2011reading} (Fig \ref{cifar_svhn}), and the comparison framework designed by Oliver \cite{oliver2018realistic} re-implemented in PyTorch \cite{paszke2017automatic}. The  comparison methods were $\Pi$-Model \cite{laine2016temporal}\cite{sajjadi2016regularization}, Mean Teacher \cite{tarvainen2017mean}, VAT \cite{miyato2018virtual}, and Pseudo-Label \cite{lee2013pseudo}, all using the Oliver framework \cite{oliver2018realistic}.

To simulate semi-supervised data, labels were removed from the majority of training data, leaving a small portion of labeled data, the size of which was systematically varied as part of the experiment. To simulate partial label information, equivalence classes were computed on the set chosen to be unlabeled, but before the labels were removed, one equivalence class per unique label value. Performance in these experiments represents an upper bound on the benefit we can expect to achieve using Nullspace Tuning over similar data, because natural partial labels are not always known so completely.

Test error and standard deviation was computed for labeled dataset sizes between 250 and 8,000, for five randomly seeded splits each. CIFAR-10 has a total of 50,000 examples of which 5,000 are set aside for validation, and SVHN a total of 73,257 of which 7325 are set aside for validation. The standard test set for each dataset are used to evaluate models. Hyperparameters for this experiment were set to those used by Oliver \cite{oliver2018realistic}.

\subsubsection*{Combined Methods} 
These experiments evaluate the benefit of adding Nullspace Tuning to an existing powerful semi-supervised learning approach. MixMatch is a good example for this demonstration, because aside from being state of the art, it uses several techniques in combination already, and therefore has a fairly complex loss function. 

Unlabeled and partially labeled examples were computed as they were for the standalone methods. For this experiment, we evaluate only on CIFAR-10, on which MixMatch has previously achieved the largest error reduction compared to other methods \cite{berthelot2019mixmatch}.  We used the TensorFlow \cite{abadi2016tensorflow} MixMatch implementation, written by the original authors \cite{berthelot2019mixmatch}, augmenting it to produce our MixMatchNST algorithm. Test error and standard deviation was computed for labeled dataset sizes between 250 and 4,000, with five random splits each.

MixMatch hyperparameters were set at the optimal CIFAR-10 settings established by Berthelot. For MixMatchNST, we set the Nullspace Tuning weight $\lambda_E=1$, which generally works well for most experiments. However, to investigate whether the addition of the Nullspace Tuning term altered the loss landscape, we performed univariate grid search over the MixUp hyperparameter $\alpha$ and the loss component weights $\lambda_E$ and $\lambda_U$ for MixMatchNST. We did not apply a linear rampup \cite{tarvainen2017mean} to $\lambda_E$ as is done for $\lambda_U$. 

Further characterization of MixMatchNST is accomplished through modifying the unlabeled data. With 500 labeled data in the CIFAR-10 training set, the model is trained with a varied amount of unlabeled data starting with all 44,500 and ending with 5,000 for one experiment. Another experiment uses all 44,500 unlabled data, but increases the chance of an nonequivalent pair provided to the Nullspace Tuning term.

Additionally, we evaluate MixMatchNST on the CIFAR-100 dataset with 10,000 labeled data using a wider model. In following the parameters used by Berthelot et al., we set $\lambda_U=150$. To find the optimal $\lambda_E$, we incrementally increase the value and retrain the model. 

\begin{figure*}[h!]
\vskip 0.2in
\begin{center}
\centerline{\includegraphics[width=\textwidth]{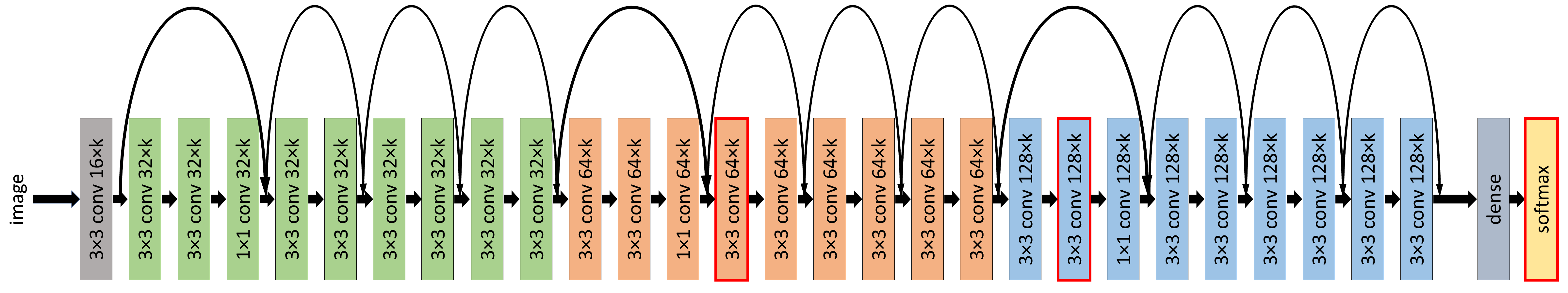}}
\caption{We choose to use a Wide ResNet-28 architecture \cite{zagoruyko2016wide} for our models. The layers highlighted with a red border are chosen to have feature maps visualized in Fig \ref{layer_vis}.}
\label{network}
\end{center}
\vskip -0.2in
\end{figure*}

To investigate how the network was responding to Nullspace Tuning, we visualized three layers in the Wide ResNet-28 model (Fig \ref{network}) for both MixMatch and MixMatchNST. The feature maps for the CIFAR-10 test set were extracted after training with 500 and 2,000 labeled examples and then were reshaped in to a vector for each sample. These flattened feature maps were then embedded in a 2D manifold fit with UMAP \cite{mcinnes2018umap} resulting in a single coordinate for each sample. 

\subsection*{Results}

\subsubsection*{Standalone Methods}
\begin{figure*}[h!]
\begin{center}
\centerline{\includegraphics[width=14cm]{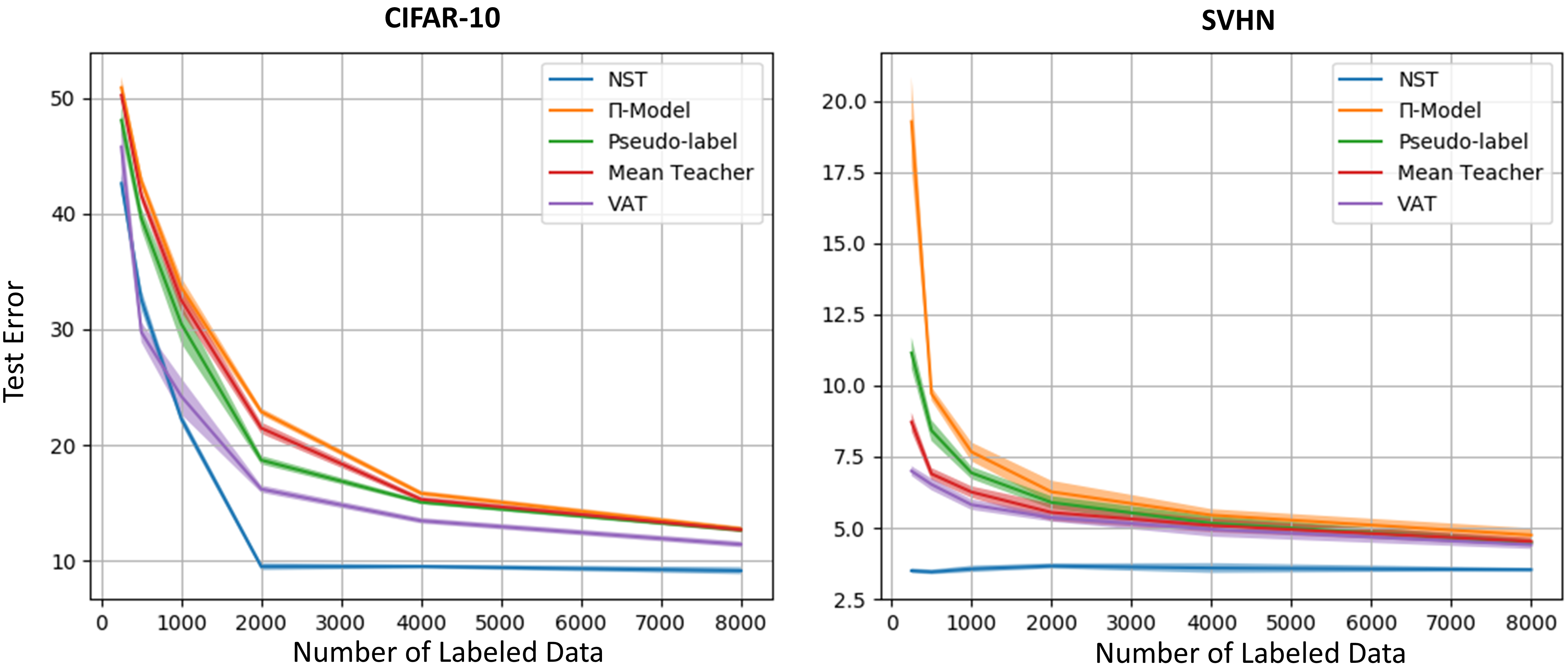}}
\caption{The added partial label information results in a substantial improvement in performance over baseline methods. This is shown in a percent test error and standard deviation (shaded region) comparison of Nullspace Tuning to baseline methods on CIFAR-10 (left) and SVHN (right) for a varied number of labeled data between 250 and 8,000. The largest improvement between Nullspace Tuning and the next best performing method (VAT) occurs in CIFAR-10 at 2,000 labeled data.}
\label{ssl_eval}
\end{center}
\vskip -0.2in
\end{figure*}

The use of partial labels generally provided a performance improvement at least as large as the difference between the best and the worst semi-supervised methods, except at the smallest labeled set sizes (Fig \ref{ssl_eval}). Surprisingly, the benefit of partial labels essentially maxes out at the relatively small number of 2,000 labeled examples (vs. 43,000 unlabeled examples) in CIFAR-10, and at less than 250 examples (vs. 65,682 unlabeled examples) in SVHN, while the semi-supervised methods continue to improve with more labeled data. 

We attribute the generally weaker performance of all methods on CIFAR-10 vs. SVHN (Fig \ref{ssl_eval}), including the large number of labeled examples needed to approach asymptotic accuracy, to the higher complexity of the images and the greater difficulty of the task (Fig \ref{cifar_svhn})

\subsubsection*{Combined Methods}
\begin{figure*}[h!]
\begin{center}
\centerline{\includegraphics[width=7cm]{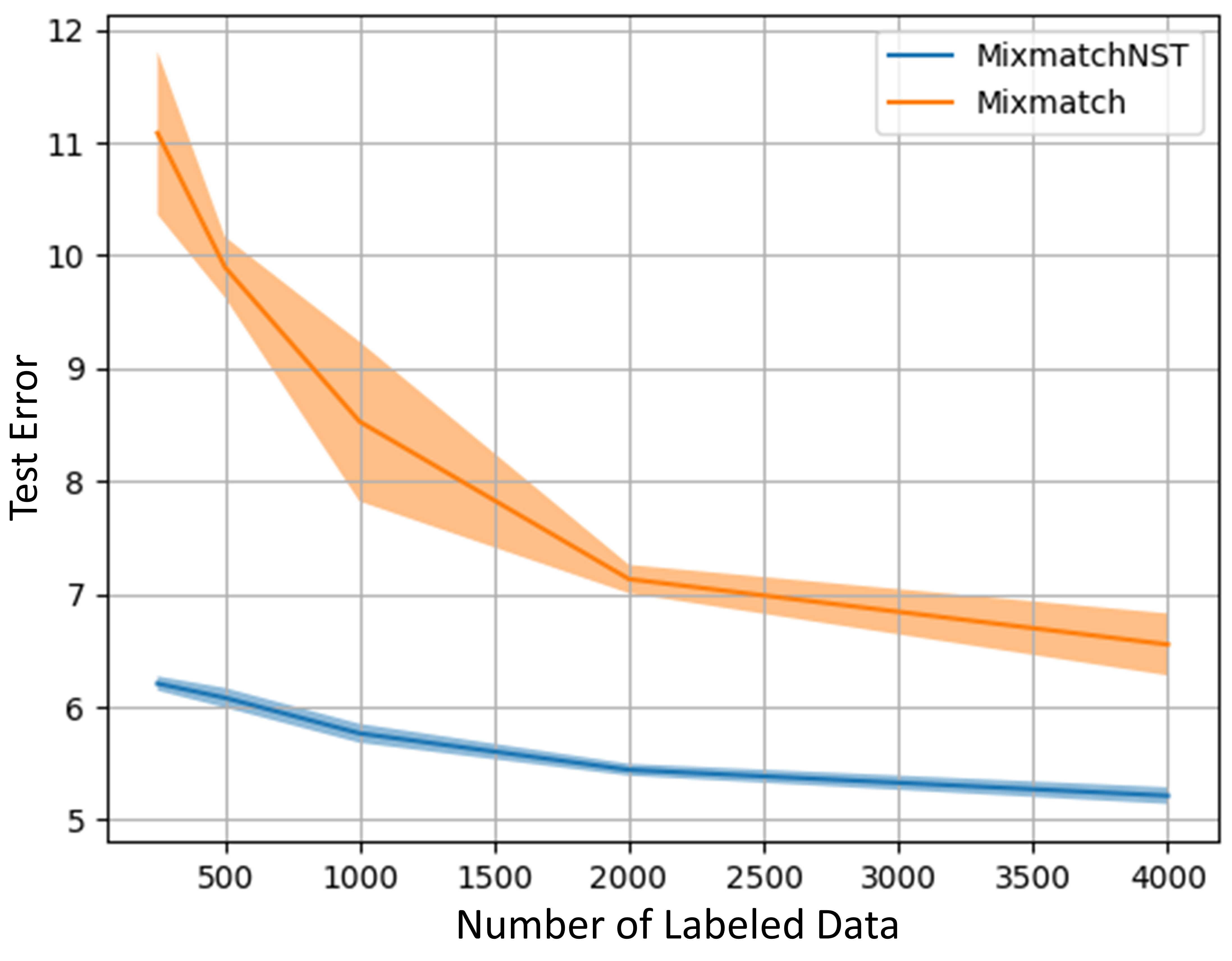}}
\caption{The additive performance of Nullspace Tuning on top of the state-of-the-art MixMatch  algorithm is considerable. This is especially evident at 250 labeled data in CIFAR-10 where error is reduced by a factor of 1.8. This is shown in a percent test error and standard deviation (shaded regions) comparison of MixMatchNST to MixMatch on CIFAR-10 for a varying number of labels.}
\label{mixmatch_eval}
\end{center}
\vskip -0.2in
\end{figure*}

The performance of MixMatch on CIFAR-10 was better than any algorithm alone, including Nullspace Tuning, in the first experiment (Fig \ref{ssl_eval}). Despite this impressive gain, performance was improved further by including Nullspace Tuning together with the MixMatch innovations. Doing so reduced test error by an additional factor of 1.8 on the smallest labeled set size, and about 1.3 at the largest set size (Fig \ref{mixmatch_eval}). 

Adding Nullspace Tuning to MixMatch with even a small number of labeled examples dramatically improved the performance on CIFAR-10 (Fig \ref{mixmatch_eval}), suggesting complementary and synergistic use of information in the two methods; either method on its own needed over 2000 labeled examples to approach its asymptotic accuracy. Test error for experiments with 250 and 2,000 labeled data are reported in Table \ref{error-table}.

\begin{table*}[h!]
\caption{CIFAR-10 and SVHN percent classification error is reported here for all methods at 250 and 2,000 labeled data. Bolded values indicate the best performing method for the number of labeled data in the dataset.}
\label{error-table}
\vskip 0.15in
\begin{center}
\begin{small}
\begin{sc}
\resizebox{14cm}{!}{\begin{tabular}{lcccr}
\toprule
method & cifar-10 error & cifar-10 error & svhn error & svhn error \\
&  250 labeled data & 2,000 labeled data &  250 labeled data &  2,000 labeled data \\
\midrule
$\Pi$-Model    & 50.88$\pm$ 0.94& 22.88$\pm$ 0.30& 19.28$\pm$ 1.58& 6.27$\pm$ 0.39 \\
Mean Teacher & 50.22$\pm$ 0.0.39& 21.46$\pm$ 0.49& 8.72$\pm$ 0.33& 5.54$\pm$ 0.30 \\
Pseudo-label    & 48.06$\pm$ 1.24& 18.70$\pm$ 0.38& 11.15$\pm$ 0.55& 5.89$\pm$ 0.22 \\
VAT    & 45.76$\pm$ 2.81& 16.19$\pm$ 0.32& 7.00$\pm$ 0.17& 5.36$\pm$ 0.14 \\
NST    & 42.60$\pm$ 0.82& 9.50$\pm$ 0.30& \textbf{3.49$\pm$ 0.08}& \textbf{3.66$\pm$ 0.10} \\
MixMatch    & 11.08$\pm$ 0.72& 7.13$\pm$ 0.13& NA & NA \\
MixMatchNST      & \textbf{6.21$\pm$ 0.06}& \textbf{5.44$\pm$ 0.05}& NA & NA \\
\bottomrule
\end{tabular}}
\end{sc}
\end{small}
\end{center}
\vskip -0.1in
\end{table*}

\newpage

\begin{figure*}[h!]
\vskip 0.2in
\begin{center}
\centerline{\includegraphics[width=14cm]{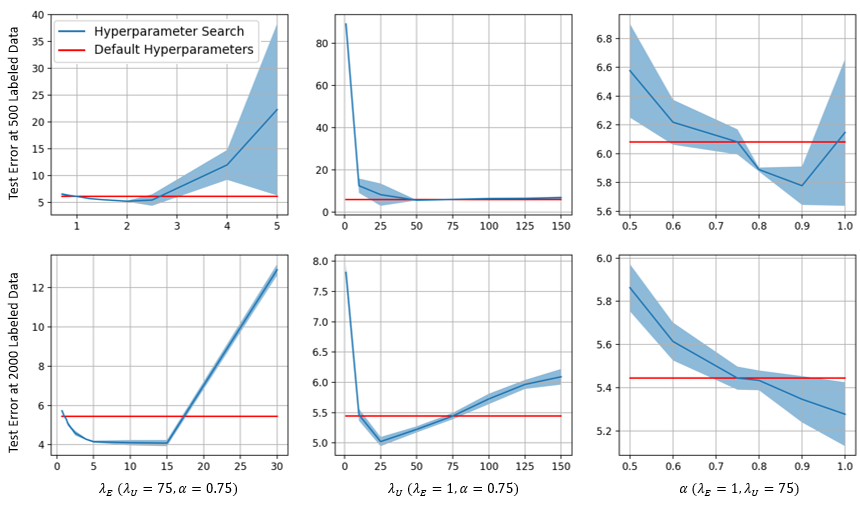}}
\caption{MixMatchNST models can benefit from hyperparameter tuning at each number of labeled data. The hyperparametr $\lambda_E$ shows the greatest need for this as the optimal value at 2,000 labeled data is at least double of that at 500 labeled data and would reduce the test error by approximately 1.4\%. Test errors and standard deviations are reported (shaded regions) at 500 labeled data (top) and 2,000 labeled data (bottom) as the hyperparameter space is searched for $\lambda_E$ (left), $\lambda_U$ (center), and $\alpha$ (right). Red lines indicate the performance before fine tuning. As one hyperparameter is tuned, the other two are set to the previously used values. The error achieved with the hyperparameters in Fig \ref{mixmatch_eval} is indicated by the red line.}
\label{hyper_search}
\end{center}
\vskip -0.2in
\end{figure*}

The hyperparameter search shows that the MixMatch loss landscape was modestly altered with respect to the MixMatch hyperparameters $\alpha$ and $\lambda_U$, and small gains could be had by tuning them further (Fig \ref{hyper_search}). Tuning our nullspace weight made a larger relative difference, providing a further improvement of about 20\% at 500 labeled datapoints, and about 30\% at 2,000 labeled datapoints, over what is shown in Fig \ref{mixmatch_eval}. 

\begin{figure*}[h!]
\vskip 0.2in
\begin{center}
\centerline{\includegraphics[width=14cm]{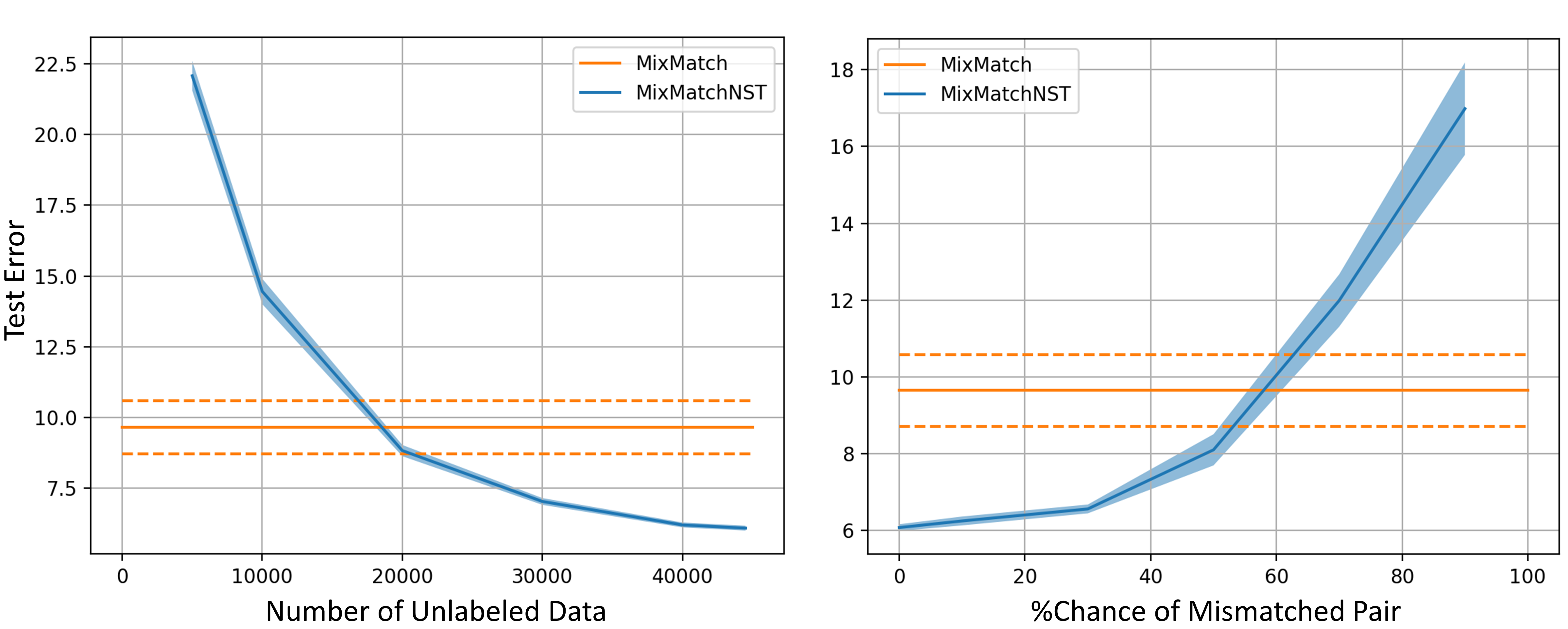}}
\caption{MixMatchNST as we alter the unlabeled data is compared to the baseline MixMatch as reported by Berthelot et al. with 500 labeled data all unlabeled data. We choose to alter the amount of unlabeled data (left) and to simulate error in the chosen equivalency classes (right). If we take away approximately half of the unlabeled data or if we have a 50\% chance of incorrectly choosing an equivalent pair, MixMatchNST still outperforms MixMatch with all unlabeled data.}
\label{ablation}
\end{center}
\vskip -0.2in
\end{figure*}

The robustness of MixMatchNST is evaluated as we altered the amount of unlabeled data and retrained the model. In reducing the amount of unlabeled data, we found that MixMatchNST can outperform the MixMatch model trained with all 44,500 unlabeled data when only 20,000 unlabeled data are available. We also found that MixMatchNST can outperform MixMatch when there is a 50\% chance that the partial label information which provides an equivalence class pair is incorrect (Fig \ref{ablation}). 

\begin{figure*}[h!]
\vskip 0.2in
\begin{center}
\centerline{\includegraphics[width=7cm]{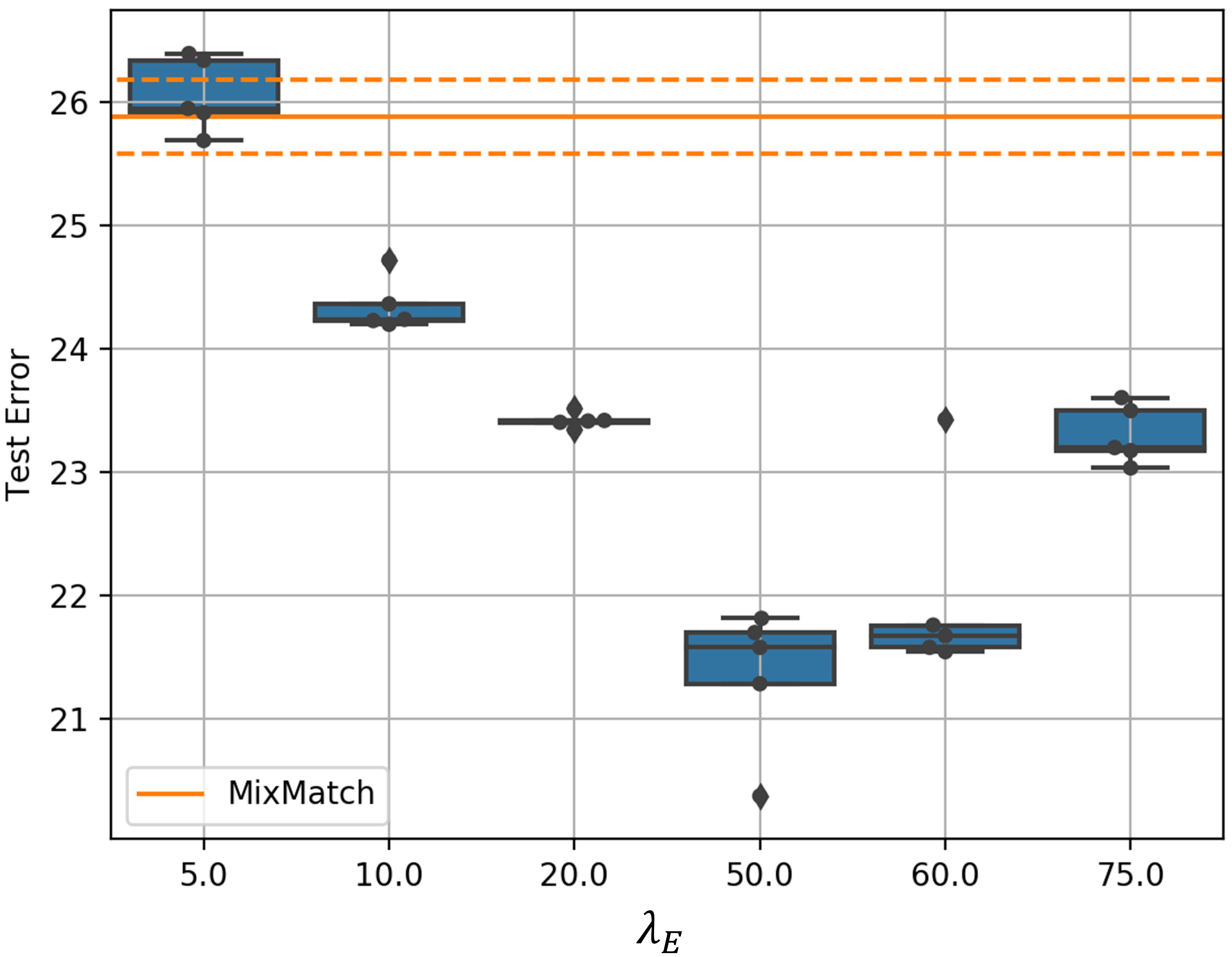}}
\caption{Using a wider network, we evaluate MixMatchNST on the CIFAR-100 dataset as we set $\lambda_U=150$ and $\alpha=0.75$ as we increase $\lambda_E$. A much larger $\lambda_E$ is needed as compared to the smaller model in the CIFAR-10 dataset.}
\label{cifar100}
\end{center}
\vskip -0.2in
\end{figure*}

MixMatchNST also sees a large increase in performance over MixMatch when evaluated on a more difficult task and using a larger model. MixMatchNST reduces test error in the CIFAR-100 dataset by more than 4\% with $\lambda_E=50$ (Fig \ref{cifar100}). 

\begin{figure*}[h!]
\begin{center}
\centerline{\includegraphics[width=\linewidth]{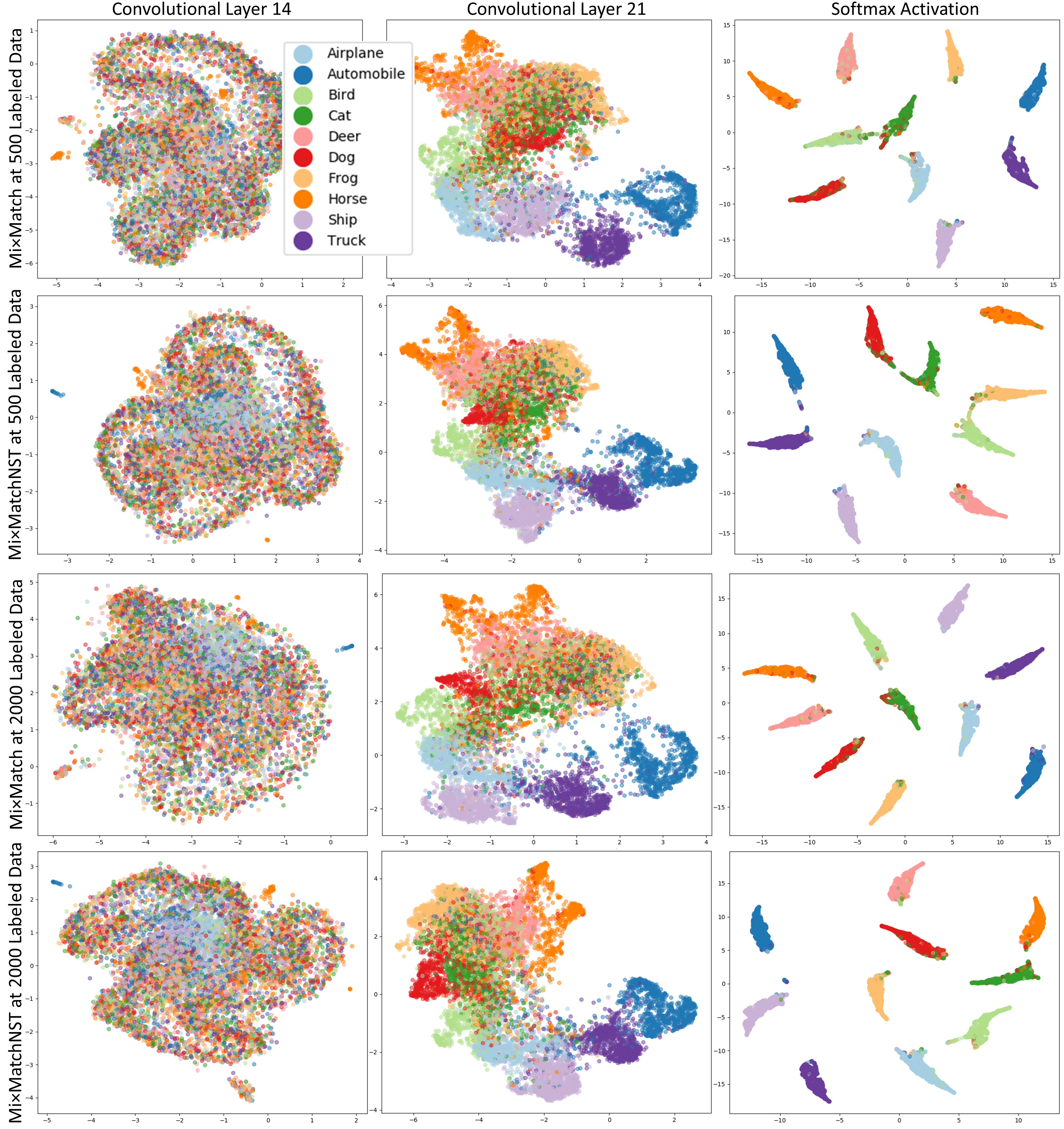}}
\caption{Nullspace Tuning provides better learning with fewer labeled examples, as evidenced by discernably clearer clusters in a 2D manifold space learned from the feature maps. In general, MixMatchNST does about as well in the CIFAR-10 test set with 500 labeled training points (second row) as MixMatch does with 2,000 (third row). Each point represents a sample's feature maps flattened to a single vector from two convolutional layers (first and second column) and the final softmax layer (third column), embedded in a 2D space learned with UMAP \cite{mcinnes2018umap}. These are shown for a single fold for MixMatch and MixMatchNST for datasets with 500 (top two rows) and 2,000 (bottom two rows) labeled examples. With 500 labeled examples, a cluster is forming at layer 14 for the class ``Airplane" in MixMatchNST, with no clear counterpart in MixMatch. At layer 21, several clusters are slightly clearer, with  separation between Cat and Dog further along. With 2,000 labeled examples, both methods are starting to form clusters for Airplane at layer 14, but MixMatchNST now also has a cluster formed for ``Ship". At layer 21, several clusters are again slightly clearer for MixMatchNST, with separation especially evident for ``Frog".}
\label{layer_vis}
\end{center}
\end{figure*}

Image features from MixMatch models and MixMatchNST models show that comparable learning happens with fewer examples with the addition of Nullspace Tuning (Fig \ref{layer_vis}), and that this learning occurs deep inside the model, rather than superficially at a later layer. The clusters in convolutional layers under Nullspace Tuning with 500 labeled examples look comparable to those for 2,000 labeled examples without it, and the clustering appears slightly clearer with Nullspace Tuning given the same number of labeled examples.  Differences in the softmax layer are subtler, but their presence is evident by the overall model performance. 

\section*{Discussion}
The main contribution of this work is the systematic demonstration that tuning the nullspace of a model using the partial label information that may reside in unlabeled data can provide a substantial performance boost compared to treating them as purely unlabeled data. It is not surprising that adding new information to a model provides such an improvement; our goal with this work was to quantify just how much improvement one could expect if equivalence classes were known within the unlabeled data. This idea is important because identifying or obtaining equivalence classes within unlabeled data may be cheaper than obtaining more labels, if standard semi-supervised methods provide insufficient performance. 

The gain from using partial label information is fairly constant over the range of labeled dataset size tested, as long as a minimum threshold of labeled data is met. This makes sense from the perspective of tuning the null space of the model, because most of that tuning can be done with equivalence classes, but a small amount of labeled data is needed to anchor what is learned to the correct labels. 

Increasing the number of labeled examples beyond the threshold is essentially trading partial label information for full label information. The relative value of that information for a given learning problem is suggested by the slope of the error curve. For the standalone methods comparison, the nearly horizontal slope suggests that partial information is nearly as good as full label information. The performance of the architecture on a fully-labeled dataset was $2.59\%$ error, which reinforces this idea. The steeper (but still mild) slope found in the combined methods comparison suggests a stronger tradeoff, although performance of MixMatch on the full $40,000$ examples is $4.2\%$ \cite{berthelot2019mixmatch}, which is fairly close to the $5.5\%$ that we get using more than $90\%$ partial labels, or even the $6.0\%$ that we get with $99\%$ partial labels. We conclude that at least in some cases, partial labels can get us most of the way there.

We can infer something about what the models are learning from the ordering of model performance: MixMatchNST $>$ MixMatch $>$ Nullspace Tuning $>$ single data-augmentation models. Explicitly learning the shape of the nullspace from partial labels was much more effective than implicitly placing data transformations into that space by the standalone algorithms, although combining those transformations into MixMatch was more effective still. But the fact that MixMatchNST performed better than either MixMatch or Nullspace Tuning alone demonstrates that MixMatch is learning somewhat different aspects of the nullspace than that provided by the partial labels.

Decreasing the amount of unlabeled data has a nonlinear effect on the performance of the MixMatchNST method, but when the amount unlabeled data is decreased by 55\%, the partial label information is able to compensate achieving better performance than a model without partial label information with all the unlabeled data. The nullspace tuning term in the MixMatchNST method is directly shown to be resilient to noise in the equivalency classes showing improvement even while 50\% of the pairs provided are not equivalent. This would suggest that even less than perfect partial labelling methods may still adequately tune the model. 

One strength of this method is its simplicity --- it can be added to nearly any other semi-supervised learning algorithm, as long as we have access to the loss function, and we can provide appropriate example pairs from an equivalence class.

This experiment used the largest possible equivalence classes --- one class for each label value. Naturally occurring equivalence classes are likely not to be so large, especially if they are obtained by repeated observations of the same object. Our experimental design investigated the most we could gain from using the partial information in equivalence classes, but if the classes are smaller and more numerous, then we might expect that gain to be smaller. But because the partial labels are given to the algorithm as example pairs, with no required relationship between those pairs and the labeled pairs, Nullspace Tuning can still be used even with equivalence classes as small as two examples. And if those equivalence classes are well distributed over the data space, their diminished size may not actually impact the benefit by much. One could imagine that even a relatively small number of relatively small equivalence classes could be rather effective at tuning the null space. The large number of trained models needed to characterize how the benefit changes with respect to the size and number of equivalence classes placed that question out of scope for this paper, but it will be an interesting direction for future work.

And of course, not all learning problems have natural equivalence classes embedded in them at all. Benchmark public datasets tend not to, except in simulations like ours, partly because information about how they were collected has been lost. But it may be cheap to instrument data collection pipelines to record information that does provide this information. In addition to the medical use cases described above, where the patient identity is tracked through repeated observations, unlabeled objects may be tracked through sequential video frames, fixed but unlabeled regions may be identified for multiple passes of a satellite, or the unlabeled sentiment of all sentences in a paragraph might be considered to form an equivalence class. We expect that there are many creative ways to find partial labels in naturally occurring datasets, and when we find them, Nullspace Tuning is a promising method to exploit them.

\section*{Conclusion}
We show that the use of equivalence classes significantly improves model performance when readily available, and while the equivalence classes are simulated, the potential increase in performance is worth obtaining the partial label information where possible. Nullspace tuning is a flexible approach that is amenable to real world learning scenarios and promises to enable use of partial label information that is not accessible with current standard neural network approaches.

\section*{Acknowledgments}
This work was submitted for review May 29, 2021. This work was supported by the National Institutes of Health under award numbers R01EB020666, R01EB017230, and T32EB001628, by the National Science Foundation under award number 1452485, and in part by the National Center for Research Resources, Grant UL1 RR024975-01. The content is solely the responsibility of the authors and does not necessarily represent the official views of the NIH. This study was in part using the resources of the Advanced Computing Center for Research and Education (ACCRE) at Vanderbilt University, Nashville, TN which is supported by NIH S10 RR031634.

\clearpage

%
%
%
\bibliography{main.bib}

\end{document}